\documentclass[10pt,twocolumn,letterpaper]{article}

\usepackage[accsupp]{axessibility}

\usepackage{cvpr}              

\usepackage{graphicx}
\usepackage{amsmath}
\usepackage{amssymb}
\usepackage{booktabs}

\usepackage{multirow}
\usepackage{bm}
\usepackage{color}
\usepackage[table,xcdraw,dvipsnames]{xcolor}
\definecolor{darkspringgreen}{rgb}{0.0, 0.45, 0.05}
\definecolor{carmine}{rgb}{0.68, 0.05, 0.0}
\definecolor{iODBlue}{RGB}{220, 232, 250}
\usepackage{pifont}
\newcommand{\cmark}{\ding{51}}%
\newcommand{\xmark}{\ding{55}}%

\usepackage[pagebackref,breaklinks,colorlinks]{hyperref}

\usepackage[capitalize]{cleveref}
\crefname{section}{Sec.}{Secs.}
\Crefname{section}{Section}{Sections}
\Crefname{table}{Table}{Tables}
\crefname{table}{Tab.}{Tabs.}

\def\cvprPaperID{6901} 
\def\confName{CVPR}
\def\confYear{2022}

\begin{document}

\title{Continual Object Detection via\\ Prototypical Task Correlation Guided Gating Mechanism}

\author{Binbin Yang\textsuperscript{1}\thanks{Equal contribution. $^\dag$Corresponding author.} \quad Xinchi Deng\textsuperscript{1}$^*$ \quad Han Shi\textsuperscript{2} \quad Changlin Li\textsuperscript{3} \quad Gengwei Zhang\textsuperscript{1} \\ \quad Hang Xu\textsuperscript{4} \quad Shen Zhao\textsuperscript{1} \quad Liang Lin\textsuperscript{1}$^\dag$ \quad Xiaodan Liang\textsuperscript{1} \\{\normalsize
\textsuperscript{1}Sun Yat-sen University \quad \textsuperscript{2}The Hong Kong University of Science and Technology} \\ {\normalsize \quad \textsuperscript{3}ReLER, AAII, UTS \quad \textsuperscript{4}Huawei Noah's Ark Lab}\\{\tt\small \{yangbb3, dengxch5\}@mail2.sysu.edu.cn, hshiac@cse.ust.hk, zhaosh35@mail.sysu.edu.cn, }\\{\tt\small \{changlinli.ai, zgwdavid, chromexbjxh, xdliang328\}@gmail.com, linliang@ieee.org}
}

\maketitle

\begin{abstract}
Continual learning is a challenging real-world problem for constructing a mature AI system when data are provided in a streaming fashion. Despite recent progress in continual classification, the researches of continual object detection are impeded by the diverse sizes and numbers of objects in each image.
Different from previous works that tune the whole network for all tasks, 
in this work, we present a simple and flexible framework for continual object detection via p\textbf{R}ot\textbf{O}typical ta\textbf{S}k corr\textbf{E}la\textbf{T}ion guided ga\textbf{T}ing mech\textbf{A}nism (ROSETTA).
Concretely, a unified framework is shared by all tasks while task-aware gates are introduced to automatically select sub-models for specific tasks. 
In this way, various knowledge can be successively memorized by storing their corresponding sub-model weights in this system. To make ROSETTA automatically determine which experience is available and useful, a prototypical task correlation guided Gating Diversity Controller (GDC) is introduced to adaptively adjust the diversity of gates for the new task based on class-specific prototypes. GDC module computes class-to-class correlation matrix to depict the cross-task correlation, and hereby activates more exclusive gates for the new task if a significant domain gap is observed. Comprehensive experiments on COCO-VOC, KITTI-Kitchen, class-incremental detection on VOC and sequential learning of 
four tasks show that ROSETTA yields state-of-the-art performance on both task-based and class-based continual object detection. \footnote{Codes are available at: \url{https://github.com/dkxocl/ROSSETA}.}
\end{abstract}

\begin{figure}[t!]
\center{
\includegraphics[width=0.45\textwidth]{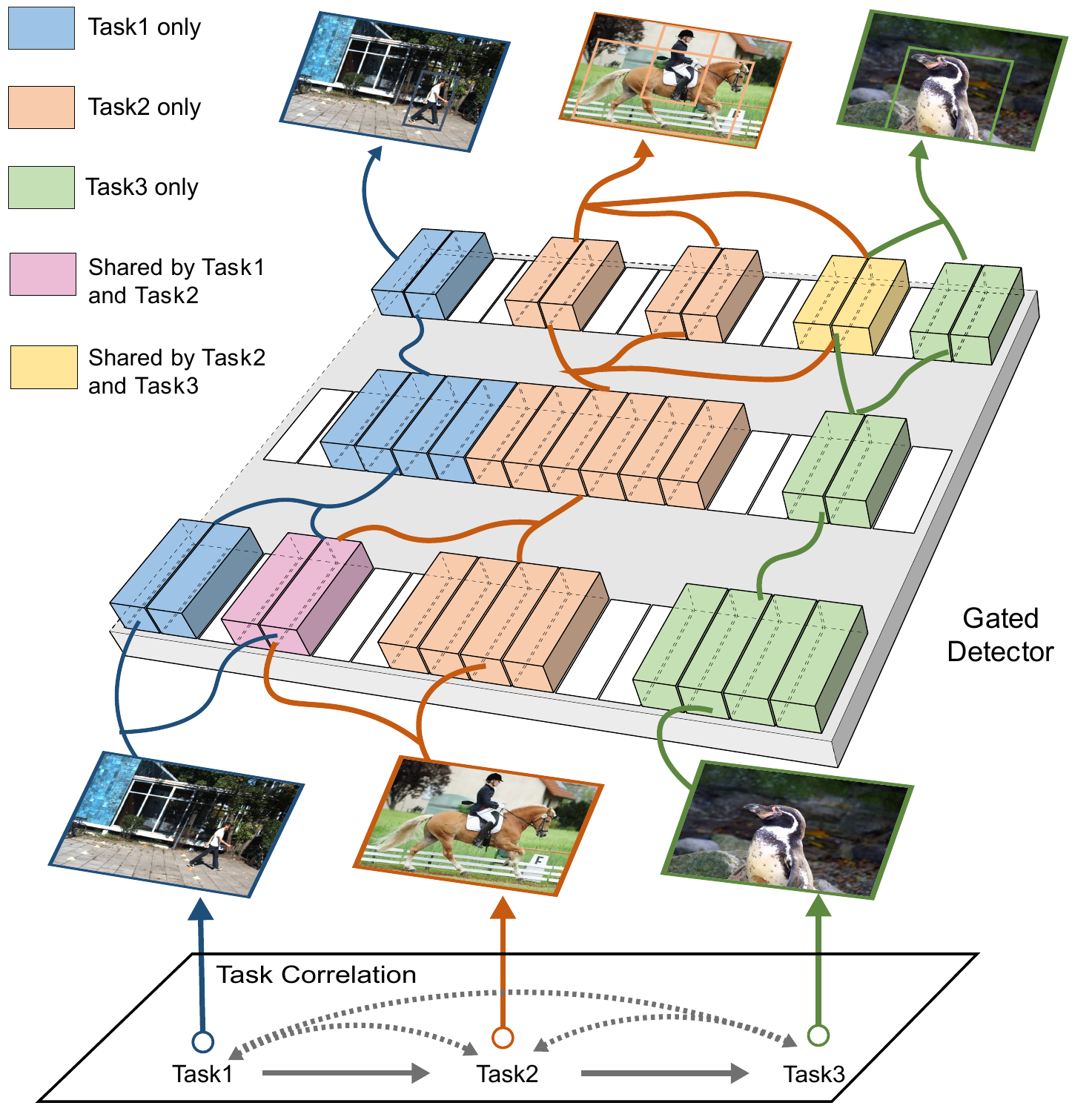}\vspace{-2mm}
  \caption{Overview of our p\textbf{R}ot\textbf{O}typical ta\textbf{S}k corr\textbf{E}la\textbf{T}ion guided ga\textbf{T}ing mech\textbf{A}nism (ROSETTA) for continual object detection. Sequential detection tasks share a unified backbone of the gated detector. Knowledge from previous tasks can be stored in the weights of the corresponding sub-models which are activated by the stored gates. Boxes in colors indicate the channels activated by task-aware gates for different tasks. Best viewed in color.\vspace{-6mm}
  }
 \label{fig:figure1}
}
\end{figure}

\vspace{-5pt}
\section{Introduction}
\vspace{-4pt}
\label{sec:intro}
Thanks to the development of computer vision and deep learning, a great progress have been made in object detection~\cite{ren2015faster, lin2017feature, lin2017focal}. The mainstream of existing works typically follows the offline training paradigm: an individual model is trained on a dataset and then evaluated on the test set with similar distribution. Nevertheless, online training on streaming data plays a more important role in real-world applications, especially large-scale industrial systems. More significantly, an artificial intelligent system is expected to continually learn different skills, \eg, detect more and more objects in multiple scenarios, resembling the memorizing and learning abilities of humans instead of learning from scratch every time. A common solution from the machine learning perspective is continual learning (lifelong learning)~\cite{kirkpatrick2017overcoming, serra2018overcoming, liu2020multi, joseph2021towards}, aiming to sequentially solve non-stationary tasks with ideally no performance drop when inferred on the previously seen tasks. Typically, an elastic model is required to achieve the equilibrium between continuously acquiring new knowledge and preserving existing knowledge.

Despite the increasing attention on continual image classification~\cite{kirkpatrick2017overcoming, serra2018overcoming, abati2020conditional}, few studies have been devoted to building a continual object detection~\cite{liu2020multi, joseph2021towards} framework due to the challenges of preserving the capability of localizing and recognizing multiple objects of diverse scales in streaming tasks. What is noteworthy is that, an image of the new task might simultaneously contain objects of novel classes and previously seen classes. 
Existing works on continual object detection mostly use knowledge distillation~\cite{liu2020multi, zhou2020lifelong, peng2020faster, shmelkov2017incremental} on features of the region proposal network (RPN) to mitigate catastrophic forgetting.
Specifically, suppose an object detector is trained consecutively on two tasks, \emph{i.e.,} $\mathtt{task1}$ and $\mathtt{task2}$. When training on $\mathtt{task2}$, the distillation-based methods will force the features of the RPN to be consistent with those produced by the saved model trained on $\mathtt{task1}$. However, such methods suffer from domain shift: distillation over the $\mathtt{task2}$ data captures biased rather than the actual knowledge of $\mathtt{task1}$ due to the unavailability of samples from $\mathtt{task1}$.
Another line of researches~\cite{joseph2021towards, liu2020multi} on continual detection store a small number of samples(exemplars) for reviewing previous knowledge. Nevertheless, replaying exemplars also fails in capturing actual knowledge because not all samples are accessible and the sampling strategy plays a decisive role.

In this work, we explore a different way to solve continual object detection without any exemplar replay: \textit{directly storing knowledge} via a sparse and dynamic framework. As illustrated in~\cref{fig:figure1}, a unified detector is shared by sequential tasks and the task-aware gates are designed to automatically determine which sub-models (channel-level) should be activated for specific tasks. To avoid the difficulty of jointly optimizing binary gates and channel weights, we propose a soften-and-discretize strategy for the gate learning. Specifically, dynamic soft gates are generated during training stage and then discretized to be static binary gates for inference stage. The channels’ weights that have been activated for previous tasks by the binary gates are frozen and stored to keep existing knowledge. Each sub-model can dynamically choose whether to use the frozen channels to boost the learning for the current task. In this manner, the previous knowledge can be unbiasedly stored in the sub-models' weights and cross-task knowledge can be shared by their overlapped channels. Binary gates 
are used to retrieve such existing knowledge by activating sub-models' channel weights.

Although the proposed gating mechanism well preserves the previous knowledge, we observe degradation on the subsequent tasks when the domain gaps are significant, \emph{e.g.}, KITTI$\rightarrow$Kitchen, which have different foreground objects and backgrounds. We attribute this phenomenon to the unawareness of cross-task correlation. Confronted with a relatively large gap, sharing too much previous knowledge would limit the performance gain in subsequent tasks~\cite{lee2021sharing} and more exclusive channels should be activated for new tasks. Thus, we propose the Prototypical Task Correlation Guided Gating Mechanism to achieve a balance between sharing existing knowledge and exploiting exclusive knowledge~(\emph{i.e.,} identifying which knowledge in the current task is orthogonal to the existing one). Specifically, a task correlation guided Gating Diversity Controller (GDC) is proposed to adaptively adjust the diversity of gates for the new task based on class-specific prototypes. GDC computes a cross-task class-to-class prototypical correlation matrix to depict the inter-task affinity and hereby activates more gates for the $\mathtt{task 2}$ when the domain gap between $\mathtt{task 1}$ and $\mathtt{task 2}$ is significant, and vice versa.

To verified its effectiveness, our proposed ROSETTA is evaluated on both task-based~\cite{liu2020multi} and class-based~\cite{joseph2021towards} continual object detection scenarios. For task-based settings, our method equipped with Faster R-CNN backbone outperforms the state-of-the-art benchmarks by 11.8 mAP and 3.4 mAP on COCO and VOC for COCO$\rightarrow$VOC, 5.8 mAP and 5.7 mAP on KITTI and Kitchen for KITTI$\rightarrow$Kitchen. As for the class-incremental detection, our method surpasses the state-of-the-art method by 2.2 mAP for the ``10+10" setting on VOC. Furthermore, comprehensive experiments demonstrate that our proposed gating mechanism successfully achieve an equilibrium between sharing knowledge and exploiting exclusive knowledge for multiple tasks by capturing their prototypical cross-task correlation.

\begin{figure*}[ht!]
\begin{center}
\includegraphics[width=0.95\linewidth]{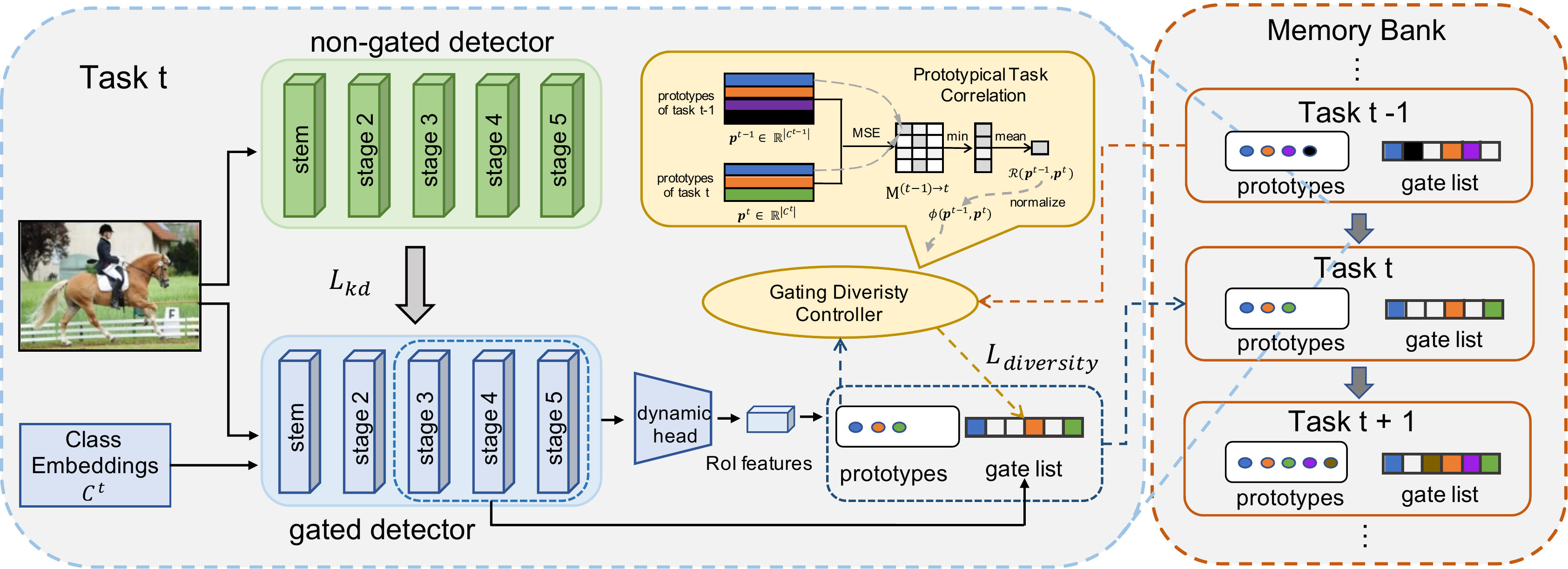}
\end{center}\vspace{-6mm}
  \caption{Pipeline of ROSETTA.~(We take the Sparse R-CNN backbone as an example in this figure.) A non-gated detector is used to guide the learning of the gated detector via feature-level knowledge distillation. Class-specific prototypes are generated as the mean ROI~(Region-of-Interest) feature of each class. A memory bank is used to store the historical gate lists and their corresponding prototypes. The gating diversity controller~(GDC) captures the prototypical cross-task correlation based on the stored gate lists and prototypes, and then automatically adjusts the diversity of gates for the current task. Best viewed in color
  \vspace{-10pt}}
\label{fig:overview}
\end{figure*}


\section{Related Work}
\label{sec:related work}
\noindent \textbf{Continual Object Detection} \; 
Continual learning~\cite{hadsell2020embracing, kirkpatrick2017overcoming, aljundi2018memory, buzzega2020dark, li2017learning} studies the problem of how an intelligent system sequentially learns from a stream of tasks. 
The ultimate goal of continual learning is to gradually digest new knowledge and preserve the acquired knowledge. Most existing studies on continual learning in visual system focus on continual image classification~\cite{chaudhry2018efficient, french1999catastrophic,kirkpatrick2017overcoming, aljundi2018selfless, riemer2018learning, lopez2017gradient} which aims to alleviate catastrophic forgetting for continually recognizing single object in an image.
In comparison, object detection \cite{ren2015faster, lin2017feature, lin2017focal} is a high-level computer vision task involving object localization and object recognition, which has not been explored totally in the continual learning scenario.
Due to the diverse sizes and numbers of objects in each image, continual object detection is a more challenging task than continual image classification. Existing works on continual object detection can be divided into two categories: storing parts of samples as exemplars for experience replay~\cite{joseph2021towards} and knowledge distillation~\cite{shmelkov2017incremental,liu2020multi}. \cite{joseph2021towards} stores a balanced set of exemplars and fine-tunes the model after each incremental step. \cite{shmelkov2017incremental} leverages knowledge distillation for both object localization and object classification. \cite{liu2020multi} further exploits attentive feature distillation to distill important knowledge via both top-down and bottom-up attentions. In this work, we explore a different way for continual object detection without using exemplars via sparse and dynamic gating mechanism.

\noindent \textbf{Gating Mechanism} \; 
Dynamic inference methods \cite{Bolukbasi2017AdaptiveNN, Huang2018MultiScaleDN,veit2018AIG,wang2018skipnet,li2020DynamicRouting} change network architecture based on the input data. As a general approach to achieve dynamic inference, gating mechanism takes in an intermediate feature map and outputs a binary vector as the decision of the candidate paths. 
It has been generally used to choose over different channels in dynamic pruning methods \cite{lin2017runtime,hua2019channel,gao2018dynamic,Chen2019YouLT, herrmann2018end,bejnordi2019batch,chen2019self} and dynamic slimmable network \cite{Li2021DynamicSN}. Such strategy has also been used in other conditional inference methods including dynamic depth \cite{veit2018AIG,wang2018skipnet,wang_dual_2020} and dynamic routing \cite{xia2021fully}.
Gating mechanism is also applied in continual image classification for dynamic inference based on the inputs from different tasks \cite{abati2020conditional, serra2018overcoming, mallya2018packnet, rajasegaran2019random}. Differing from these works that rely on gradient estimation or other techniques for optimization of the binary gate, we use dynamic soft gates during training time and then discretize to static binary ones for inference. 

\section{Methodology}
\label{sec:methodology}
\vspace{-4pt}

\subsection{Continual Object Detection and its Challenges}

\noindent \textbf{Continual Object Detection} \; The goal of continual object detection is to obtain an object detector performing well on $\mathcal{T}$ sequential Tasks. For the $t^{th}$ task, the dataset $D^t = \{X^t, Y^t\}$ is provided to train the object detector, where $X^t$ and $Y^t$ denote the input images and the corresponding annotations, respectively. The categories to be recognized at time $t$ are denoted as a set $C^t$.

\noindent \textbf{Problem of Catastrophic Forgetting} \; Catastrophic forgetting is a phenomenon that models suffer from rapid performance degradation on previously learned tasks~\cite{hadsell2020embracing}. This usually occurs when a model is continuously transferred among multiple datasets when the old data in history can not be accessed for review. Lots of works have explored to mitigate catastrophic forgetting for the problem of continual image classification, while a more delicate strategy is required for continual object detection due to the variety of numbers, sizes and classes of the objects of interests in different datasets~\cite{liu2020multi}.

In this section, we propose a new method for continual object detection to better tackle the problem of catastrophic forgetting via p\textbf{R}ot\textbf{O}typical ta\textbf{S}k corr\textbf{E}la\textbf{T}ion guided ga\textbf{T}ing mech\textbf{A}nism (ROSETTA). As illustrated in~\cref{fig:figure1} and \cref{fig:overview}, ROSETTA can memorize multiple sequentially learned knowledge by storing the weights of one unified object detector, without any exemplar replay. For convenience of computing cross-task correlation, a memory bank is used to store each task's gate list and class-specific prototypes. Thanks to the gating strategy, knowledge of the previously seen tasks can be queried by their corresponding gates and retrieved by activating sub-models' channel weights. Moreover, a task correlation guided Gating Diversity Controller (GDC) is introduced to capture the cross-task correlation according to the stored gate lists and prototypes in memory banks and then make ROSETTA dynamically determine the gating diversity. In the following parts, elements of our proposed ROSETTA will be elaborated.

\subsection{Task-aware Gated Object Detector}
\label{sec:gated_objecter}
\vspace{-4pt}

\begin{figure}[t]
\begin{center}
\includegraphics[width=0.9\linewidth]{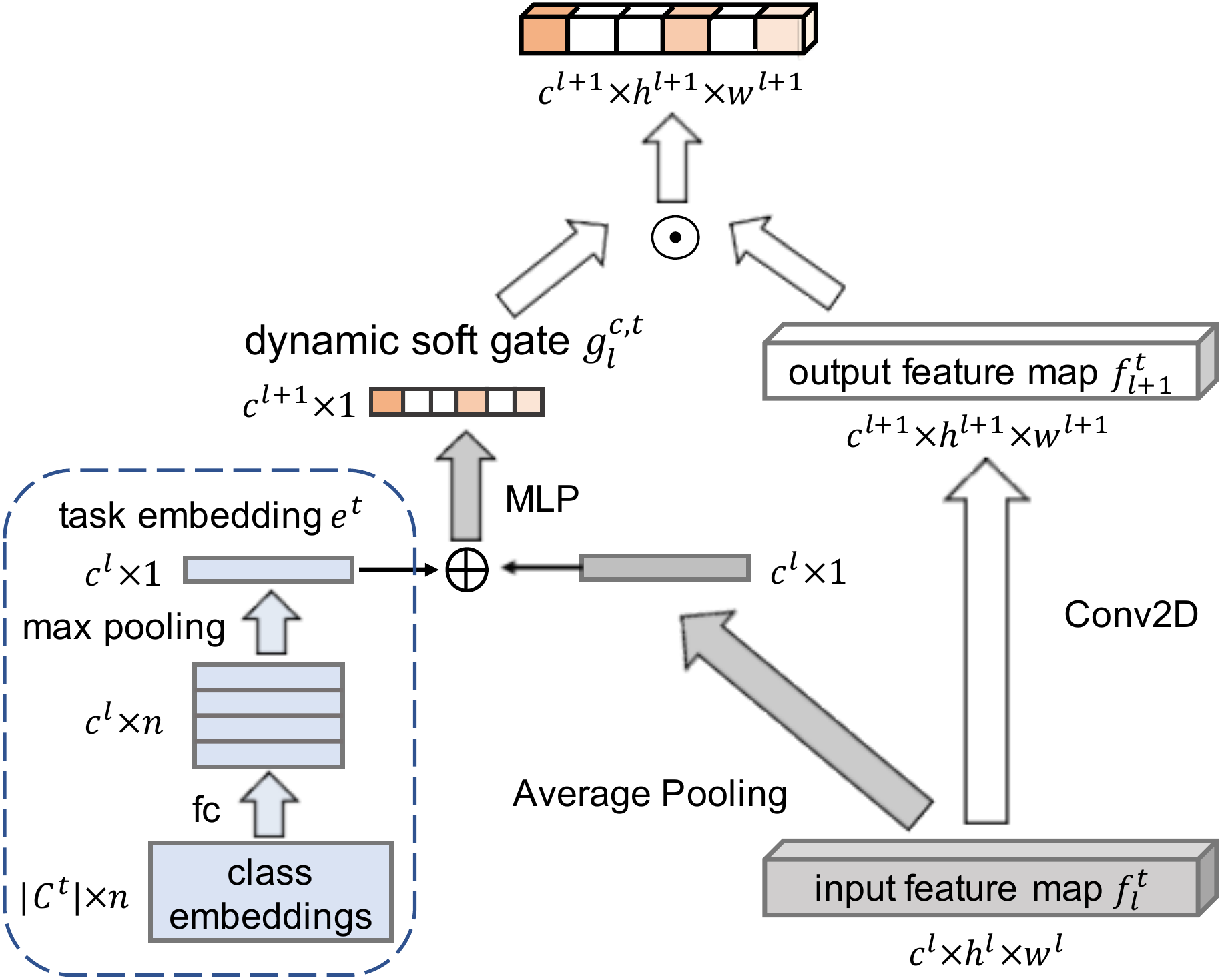}
\end{center}\vspace{-5mm}
   \caption{Our proposed gated convolutional module (training stage). Class embeddings~(word embeddings of the class labels) of the current task is fed into a fully connected layer and a max pooling layer to obtain task embedding $\bm{e}^t$. An MLP takes its inputs as task embedding $\bm{e}^t$ and input feature map $\bm{f}_l^t$ to generate the dynamic soft gate $g_l^{c,t}$.\vspace{-15pt}}
\label{fig:gate_mechanism}
\end{figure}

As illustrated in~\cref{fig:gate_mechanism}, our proposed gating mechanism is applied to the convolutional layers of an object detector, which activates the sub-model of a specific task.

Given an input feature map $\bm{f}^t_l \in \mathbb{R}^{c^l\times h^l\times w^l}$ of the $l^{th}$ layer for task $t$, a convolutional operation $F^t_l$ is performed and a gating module $G_l$ generates the activated channels:
\vspace{-5pt}\begin{equation}
\vspace{-4pt}
    \bm{f}^t_{l+1} = G^t_l(\bm{f}^t_l, C^t)\odot(F^t_l(\bm{f}^t_l)),
\end{equation}
where $\odot$ denotes the operation of channel-wise multiplication, $\bm{f}^t_{l+1} \in \mathbb{R}^{c^{l+1}\times h^{l+1}\times w^{l+1}}$ is the output feature map of gated convolution, $F^t_l(\bm{f}^t_l) \in \mathbb{R}^{c^{l+1}\times h^{l+1}\times w^{l+1}}$ is the output of convolution, $G^t_l(\bm{f}^t_l, C^t) = [g_l^{1,t}, g_l^{2,t}, ..., g^{c^{l+1, t}}_l]$ represents channel gates of the $l^{th}$ layer, $g_l^{c,t}\in[0,1]$. The learnable task embedding $\bm{e^t}$ is generated based on class embeddings $C^t$, which is the concatenation of word embeddings~\cite{mikolov2013efficient} of the classes of current task. By introducing class embeddings $C^t$, the gated module can know exactly which categories of objects need to locate and recognize. Such meta-information provides a global contextual guidance for the gated module and makes it generate different gates when two tasks have similar images (input feature maps $\bm{f}^t_l$) but different classes to be detected, \eg, class-incremental detection on VOC.

Previous works on gating mechanism~\cite{abati2020conditional,Li2021DynamicSN, serra2018overcoming, mallya2018packnet} employ binary gates,~\emph{i.e.}~$g_l^{c,t}\in\{0,1\}$, to control the selection of sub-models in both training and inference stages, which results in the difficulty of back-propagation. Empirically we found that training with binary gates easily collapses on complex and large-scale detection tasks. Inspired by the widely used continuous relaxation strategy in neural architecture search~(NAS)~\cite{liu2018darts, xu2019pc, chen2019progressive}, we propose a \textbf{soften-and-discretize} strategy for our gate learning to overcome the convergence hardship. Concretely, \textbf{dynamic soft gates} are generated by our gated module during training stage and then discretized to be \textbf{static binary gates} for inference stage via choosing appropriate thresholds $\gamma_l^{c,t}$:
\vspace{-5pt}\begin{equation}
    \gamma_l^{c,t} = \mathbb{E}_{(x,y)\sim D_\textit{val}^t}[g_l^{c,t}],
    \vspace{-4pt}
\end{equation}
where $D_\textit{val}^t$ is the validation set of the $t^{th}$ task, $c$ denotes the $c^{th}$ channel of the $l^{th}$ layer and $g_l^c$ is the corresponding channel gate mentioned above.

The static binary gates are obtained by thresholding as follows:
\vspace{-5pt}\begin{equation}
    \hat{g}_l^{c, t} = \mathbb{I}[g_l^{c,t} \geq \gamma_l^{c,t}],
\end{equation}
where $\mathbb{I}[\cdot]$ is an indicator function. It should be noticed that the dynamic soft gates are input-dependent while the static binary gates are \emph{input-independent} and applicable to all samples in a task in inference stage. Similar to the coupling problem of discrete architecture encoding and the sub-model weights~\cite{liu2018darts,guo2020single} in NAS~(retraining or fine-tuning weights is needed), the activated channel weights and the discretized binary gates may not well accommodate to each other. To address this issue, we slightly fine-tune the newly activated channel weights to adapt it to the binary gates. Ultimately, the gated output feature for inference time is given as:
\vspace{-5pt}\begin{equation}
    \bm{f}^t_{l+1} = \hat{\bm{g}}_l^{t} \odot(F^t_l(\bm{f}^t_l)),
    \vspace{-4pt}
\end{equation}
where $\hat{\bm{g}}_l^{t} = [\hat{g}_l^{1,t}, \hat{g}_l^{2,t}, ..., \hat{g}^{c^{l+1, t}}_l]$ are binary gates.

In addition to the soften-and-discretize strategy, \textbf{sparsity constraint}~\cite{abati2020conditional} and \textbf{knowledge distillation}~\cite{hinton2015distilling} are incorporated to guide the learning of our sparse gated model. The sparsity loss encourages a smaller size of the sub-model in each task and reserves more inactivated channels for future tasks:
\vspace{-5pt}\begin{equation}
    \mathcal{L}_{\textit{sparsity}} = \mathbb{E}_{(x,y)\sim D^t}\left[\frac{1}{L}\sum_{l=1}^L\frac{\Vert G_l(\bm{f}^t_l, C^t)\Vert_1}{c^{l+1}}\right],
    \vspace{-4pt}
\end{equation}
where $L$ is the number of layers. Distinct from existing distillation-based continual learning methods~\cite{liu2020multi, zhou2020lifelong, peng2020faster, shmelkov2017incremental}, we use feature-level knowledge distillation to avoid the training collapse of the sparse gated model rather than alleviating catastrophic forgetting. As shown in~\cref{fig:overview}, the distillation loss is leveraged to let the gated student model generate similar feature maps to the non-gated teacher model:
\vspace{-5pt}\begin{equation}
    \mathcal{L}_{\textit{kd}} = \frac{1}{L}\sum_{l=1}^L\text{MSE}(\bm{f}^t_{l}, \bm{\tilde{f}}^t_{l}),
    \vspace{-4pt}
\end{equation}
where \text{MSE} is the mean square error between feature maps from student and teacher models, $\bm{\tilde{f}}^t_{l}$ is the feature map of teacher model,~\emph{i.e.} $\bm{\tilde{f}}^t_{l} = \tilde{F}^t_{l-1}(\bm{\tilde{f}}^t_{l-1})$, $\tilde{F}^t_{l-1}$ is a traditional convolutional operation without gated module.

\subsection{Task Correlation Guided Gating Diversity}
\label{sec:gating_diversity}
\vspace{-4pt}

With the help of the aforementioned task-aware gated module, existing knowledge can be fully stored in sub-models' weights and easily retrieved by their binary gates. When we move to solve the subsequent tasks, the previously activated channel weights will be frozen and the other inactivated ones will be still learnable. The sub-model for a new task can select parts of frozen channel weights of previous tasks to facilitate and promote its learning process, which can be seen as a scheme of knowledge sharing among tasks. In this way, the problem of catastrophic forgetting in continual object detection can be solved by the gating mechanism proposed in~\cref{sec:gated_objecter}. However, we observe performance degradation when learning on the subsequent tasks if domain gaps are significant. We find knowledge sharing provided by gating mechanism benefits the subsequent task if the domain gap is small, \eg,~COCO$\rightarrow$VOC. But it limits the performance gain when tasks have obviously different foreground objects and backgrounds, \eg,~KITTI is an outdoor autonomous driving dataset and Kitchen contains indoor kitchen scenes. We attribute this phenomenon to the unawareness of cross-task correlation and propose the Prototypical Task Correlation Gating Mechanism to achieve dynamic balance between sharing existing knowledge and exploiting exclusive knowledge~(\emph{i.e.} identifying the exclusive knowledge compared to the existing one).

To further exploit the potential of inactivated channels for a new task and enhance the ability to digest new knowledge, we introduce a \textbf{gating diversity loss} to allow more channel gates to be activated during training the new task. The diversity loss $\mathcal{L}_{\textit{diversity}}$ is in the form of entropy:
\vspace{-5pt}\begin{equation}
\label{eq:diversity}
\mathcal{L}_{\textit{diversity}}^{l,t} = q_{l}^t\log q_{l}^t + (1 - q_{l}^t)\log (1-q_{l}^t),
\vspace{-5pt}
\end{equation}
\begin{equation}\label{eq:openratio}
q_{l}^t = \frac{\sum_{i=1}^{m_l^t}g_l^{c,t}\mathbb{I}[g_l^{c,t}\geq \eta]}{\sum_{i=1}^{m_l^t} g_l^{c,t}},
\end{equation}
where $m_l^t$ is the total number of previously inactivated channels in layer $l$ reserved for task $t$, $\mathbb{I}$ is an indicator function, $\eta$ is a hyper-parameter threshold. $q_l^t$ in~\cref{eq:openratio} is an estimated ratio of the newly activated gates for task $t$ in layer $l$. The gating diversity loss in~\cref{eq:diversity} is a negative entropy corresponding to such estimated ratio $q_l^t$. Minimizing the gating diversity loss means more gates will be activated if necessary while the saturation of channels is avoided thanks to the property of entropy.

Taking the correlation between different tasks into account, it can be intuitively assumed that few exclusive knowledge (newly activated gates) are necessary when two tasks are similar or almost identical, and vice versa. Going one step further, our ROSETTA is expected to automatically determine whether more gating diversity is needed for a new task. Hence, a task correlation guided \textbf{Gating Diversity Controller} (GDC) is integrated to adaptively adjust the gating diversity based on class-specific prototypes. For task $t$, the prototype of the $i^{th}$ class $\bm{p}_i^t$ is stored in the memory bank as the mean RoI (Region-of-Interest) feature of class $i$, which is generated by an object detector. The class-to-class prototypical correlation matrix $\mathbf{M}^{m\rightarrow n} \in \mathbb{R}^{|C^m|\times |C^n|}$ can be used to depict the class-to-class correlation~(distance) between task $m$ and task $n$ ($m\leq n$):
\vspace{-5pt}
\begin{equation}
\label{eq:class-to-class}
\mathbf{M}^{m\rightarrow n}_{i, j} = \text{MSE}(\bm{p}^m_i, ~\bm{p}^n_j), ~i\in C^{m} \wedge j\in C^{n},
\vspace{-4pt}
\end{equation}
where $C^t$ denotes the categories to be recognized for task $t$. Based on $\mathbf{M}^{m\rightarrow n}$, we can define the class-to-task correlation between the $j^{th}$ class of task $n$ and the $m^{th}$ task ($\bm{p}^m = [p_1^m, p_2^m,...,p^m_{C^m}]$) as:
\vspace{-5pt}
\begin{equation}
\label{eq:class-to-task}
\mathcal{R}(\bm{p}^n_j, \bm{p}^m)=
\begin{cases}
 \;\;\mathop {\min }\limits_{i\in C^m} \quad \mathbf{M}^{m\rightarrow n}_{i, j}, & m < n\\
 \mathop {\min }\limits_{i\in C^m, i \ne j} \mathbf{M}^{m\rightarrow n}_{i, j}, & m = n.
\end{cases}
\vspace{-4pt}
\end{equation}
The task-to-task correlation between task $m$ and task $n$ can be induced by the average of the class-to-task correlations:
\vspace{-5pt}
\begin{equation}
\label{eq:task-to-task}
\mathcal{R}(\bm{p}^n, \bm{p}^m) = \frac{1}{C^n}\sum_{j\in C^n}\mathcal{R}(\bm{p}^n_j, \bm{p}^m), m\leq n.
\vspace{-4pt}
\end{equation}

Based on the prototypical task correlation $\mathcal{R}(\bm{p}^n, \bm{p}^m)$, the gating diversity controller~(GDC) is used to give a weight for the diversity loss in~\cref{eq:diversity} and maintain an equilibrium between sharing knowledge and exploiting exclusive knowledge:
\vspace{-5pt}
\begin{equation}
\label{eq:controller}
\phi(\bm{p}^n, \bm{p}^m) = \max\{\frac{\mathcal{R}(\bm{p}^n, \bm{p}^m) - \mathcal{R}(\bm{p}^m, \bm{p}^m)}{\mathcal{R}(\bm{p}^n, \bm{p}^m)}, 0\}.
\vspace{-4pt}
\end{equation}
The controller in~\cref{eq:controller} provides a weight in the range of $[0,1]$, which is a greater value when the knowledge transferring from task $m$ to task $n$ is more difficult and then more exclusive channels for task $n$ should be activated.

By combining \cref{eq:diversity} and~\cref{eq:controller}, the cross-task correlation guided gating diversity loss for task $t$ ($t>1$) is:
\vspace{-5pt}
\begin{equation}
\label{eq:final_diversity}
\mathcal{L}_{\textit{diversity}} = \frac{1}{L}\frac{1}{(t-1)}\sum_{l=1}^L\sum_{i=1}^{t-1}\phi(\bm{p}^t, \bm{p}^i)\mathcal{L}_{\textit{diversity}}^{l,t}.
\vspace{-5pt}
\end{equation}

To conclude, incorporating the cross-task correlation guided gating diversity loss into the continual learning pipeline of our task-aware gated object detector can achieve better equilibrium between acquiring new knowledge and preserving old knowledge by adaptively adjusting the diversity of channel gates.

\section{Experiments}
\vspace{-4pt}
\label{sec:experiments}
\subsection{Experimental Setting}
\vspace{-4pt}
\noindent\textbf{Datasets} \; In general, there are two kinds of experimental settings for continual detection: task-incremental object detection~\cite{liu2020multi} and class-incremental object detection~\cite{joseph2021towards}. In terms of task-incremental object detection, $D^i$ and $D^j$ ($i\neq j$) are two independent datasets. As for class-incremental object detection, the categories of objects we expect to detect is incrementally observed in a dataset, \emph{i.e.} $C^t \subset C^{t+1}$. In brief, the former setting is dataset-level incremental detection and the latter one is in a class-level fashion. Without loss of generality, we will evaluate our model for both these two case of continual object detection. Following the continual object detection settings in~\cite{liu2020multi} and \cite{joseph2021towards}, we conduct experiments on COCO~\cite{lin2014microsoft}, Pascal VOC~\cite{everingham2010pascal}, KITTI~\cite{geiger2012we} and Kitchen~\cite{georgakis2016multiview}. For task-incremental object detection, the datasets we use are COCO-VOC and KITTI-Kitchen, following the settings in~\cite{liu2020multi}. COCO and VOC have similar domains, and 20 categories of these two datasets are coincident. KITTI is an outdoor autonomous driving dataset while Kitchen contains indoor kitchen scenes. Therefore, domains of KITTI and Kitchen are significantly different and their categories are totally disjoint, which makes continual object detection on KITTI-Kitchen more challenging than COCO-VOC. For class-incremental continual object detection, ROSETTA is validated following the incremental protocol in~\cite{joseph2021towards}.  The datasets for incremental detection come from Pascal VOC 2007~\cite{everingham2010pascal} and three settings~(``10 + 10", ``15 + 5", ``19 + 1") are used for evaluation. We split Pascal VOC into two tasks as an incremental task sequence. For example, ``10 + 10" means that a detector is firstly trained on the images only with the annotations of the first 10 categories, and then trained on the those containing the other 10 classes.

\begin{table}[t!]
 \scriptsize
  \centering
  \setlength{\tabcolsep}{4pt}{
  \begin{tabular}{l|c|c}
    \toprule
    \multicolumn{3}{c}{COCO-VOC}\\
    \midrule
    Methods & COCO $\rightarrow$ VOC & VOC $\rightarrow$ COCO \\
    \midrule
    Joint Training (Faster R-CNN) & 48.8 \qquad 81.6 & 81.6 \qquad 48.8\\
    Fine-tuning (Faster R-CNN) & 23.2 \qquad 79.5 & 74.7 \qquad 47.1 \\
    
    \midrule
    LwF Detection$^\dag$~\cite{shmelkov2017incremental}& 26.6 \qquad 73.0 & - \\
    Feature Distillation$^\dag$~\cite{romero2014fitnets} & 26.9 \qquad 72.4 & - \\
    Attention Distillation$^\dag$~\cite{zagoruyko2016paying} & 28.5 \qquad 73.0 & - \\
    EWC$^\dag$~\cite{kirkpatrick2017overcoming}& 32.2 \qquad 73.4 & - \\
    MAS$^\dag$~\cite{aljundi2018memory} & 32.7 \qquad 73.4 & - \\
    AFD$^\dag$~\cite{liu2020multi} & 36.8 \qquad 75.2 & -  \\

    \midrule
    EWC~\cite{kirkpatrick2017overcoming} & 27.2 \qquad 75.0 & 67.0 \qquad 44.4 \\
    MAS~\cite{aljundi2018memory} & 28.1 \qquad 74.8 & 69.0 \qquad 43.9 \\
    AFD~\cite{liu2020multi} & 27.8 \qquad 77.1 & 75.4 \qquad 45.1 \\
    ROSETTA-Faster R-CNN(Ours) & \textbf{48.6} \qquad \textbf{80.5} & \textbf{77.5} \qquad \textbf{46.5} \\
    
    \midrule\midrule
    Joint Training (Sparse R-CNN) & 52.9 \qquad 83.2 & 83.2 \qquad 52.9 \\
    Fine-tuning (Sparse R-CNN) & 35.7 \qquad 81.2 & 74.9 \qquad 49.9 \\
    
    \midrule
    ROSETTA-Sparse R-CNN(Ours) & 49.5 \qquad 82.3 & 79.5 \qquad 48.3 \\
    \bottomrule
    \toprule
    \multicolumn{3}{c}{KITTI-Kitchen}\\
    \midrule
    Methods & KITTI $\rightarrow$ Kitchen & Kitchen $\rightarrow$ KITTI \\
    \midrule
    Joint Training (Faster R-CNN) &  55.0 \qquad 83.7 & 83.7 \qquad 55.0\\
    Fine-tuning (Faster R-CNN) & 7.7 \;\,\qquad 82.2 & 13.5 \qquad 54.2\\
    
    \midrule
    LwF Detection$^\dag$~\cite{shmelkov2017incremental}& 39.4 \qquad 69.9 & 59.9 \qquad 54.7\\
    Feature Distillation$^\dag$~\cite{romero2014fitnets} & 35.0 \qquad 69.4 & 62.7 \qquad 54.4\\
    Attention Distillation$^\dag$~\cite{zagoruyko2016paying} & 39.8 \qquad 71.0 & 64.2 \qquad 52.8\\
    EWC$^\dag$~\cite{kirkpatrick2017overcoming}& 48.3 \qquad 65.5 & 68.4 \qquad 52.8\\
    MAS$^\dag$~\cite{aljundi2018memory} & 42.8 \qquad 71.7 & 67.7 \qquad \textbf{55.6}\\
    AFD$^\dag$~\cite{liu2020multi} & 48.1 \qquad 72.4 & 68.6 \qquad 53.4 \\
    
    \midrule
    EWC~\cite{kirkpatrick2017overcoming} & 15.8 \qquad 65.8 & 10.7 \qquad 53.4\\
    MAS~\cite{aljundi2018memory} & 8.9 \;\,\qquad 70.3 & 11.5 \qquad 54.0\\
    AFD~\cite{liu2020multi} & 36.6 \qquad 72.0 & 20.5 \qquad 50.5\\
    ROSETTA-Faster R-CNN(Ours) & \textbf{53.9} \qquad \textbf{78.1} & \textbf{78.4} \qquad 54.7\\
    
    \midrule\midrule
    Joint Training (Sparse R-CNN) & 55.5 \qquad 81.8 & 81.8 \qquad 55.5\\
    Fine-tuning (Sparse R-CNN) &  20.2 \qquad 79.8 & 18.6 \qquad 53.6\\
    
    \midrule
    ROSETTA-Sparse R-CNN(Ours) & 53.3 \qquad 78.3 & 78.2 \qquad 54.5\\
    \bottomrule
  \end{tabular}
  \vspace{-3mm}
  \caption{Comparisons with existing methods for task-incremental object detection, in terms of mAP (\%). Arrows indicate the order of learning. Methods storing exemplars for experience replay are denoted by `$\dag$'. `-' indicates that results are not given by~\cite{liu2020multi} and no public codes are provided. Results of ROSETTA equipped with Faster R-CNN and Sparse R-CNN backbones are both shown in this table. The best results with Faster R-CNN backbone are denoted in boldface. \vspace{-15pt}}
    \label{tab:task_incre}}
\end{table}

\begin{table*}[ht!]
\centering\setlength{\tabcolsep}{3pt}
\resizebox{\textwidth}{!}{%
\begin{tabular}{@{}l|cccccccccccccccccccc|c@{}}
\toprule
{\color[HTML]{009901} \textbf{10 + 10 setting}} & aero & cycle & bird & boat & bottle & bus & car & cat & chair & cow & \cellcolor[HTML]{DAE8FC}table & \cellcolor[HTML]{DAE8FC}dog & \cellcolor[HTML]{DAE8FC}horse & \cellcolor[HTML]{DAE8FC}bike & \cellcolor[HTML]{DAE8FC}person & \cellcolor[HTML]{DAE8FC}plant & \cellcolor[HTML]{DAE8FC}sheep & \cellcolor[HTML]{DAE8FC}sofa & \cellcolor[HTML]{DAE8FC}train & \cellcolor[HTML]{DAE8FC}tv & \textbf{mAP} \\ \midrule
All 20 & 68.5 & 77.2 & 74.2 & 55.6 & 59.7 & 76.5 & 83.1 & 81.5 & 52.1 & 79.8 & \cellcolor[HTML]{DAE8FC}55.1 & \cellcolor[HTML]{DAE8FC}80.9 & \cellcolor[HTML]{DAE8FC}80.1 & \cellcolor[HTML]{DAE8FC}76.8 & \cellcolor[HTML]{DAE8FC}80.5 & \cellcolor[HTML]{DAE8FC}47.1 & \cellcolor[HTML]{DAE8FC}73.1 & \cellcolor[HTML]{DAE8FC}61.2 & \cellcolor[HTML]{DAE8FC}76.9 & \cellcolor[HTML]{DAE8FC}70.3 & 70.51\\
First 10 & 79.3 & 79.7 & 70.2 & 56.4 & 62.4 & 79.6 & 88.6 & 76.6 & 50.1 & 68.9 & \cellcolor[HTML]{DAE8FC}0 & \cellcolor[HTML]{DAE8FC}0 & \cellcolor[HTML]{DAE8FC}0 & \cellcolor[HTML]{DAE8FC}0 & \cellcolor[HTML]{DAE8FC}0 & \cellcolor[HTML]{DAE8FC}0 & \cellcolor[HTML]{DAE8FC}0 & \cellcolor[HTML]{DAE8FC}0 & \cellcolor[HTML]{DAE8FC}0 & \cellcolor[HTML]{DAE8FC}0 & 35.59 \\
New 10 & 7.9 & 0.3 & 5.1 & 3.4 & 0 & 0 & 0.2 & 2.3 & 0.1 & 3.3 & \cellcolor[HTML]{DAE8FC}65 & \cellcolor[HTML]{DAE8FC}69.3 & \cellcolor[HTML]{DAE8FC}81.3 & \cellcolor[HTML]{DAE8FC}76.4 & \cellcolor[HTML]{DAE8FC}83.1 & \cellcolor[HTML]{DAE8FC}47.2 & \cellcolor[HTML]{DAE8FC}67.1 & \cellcolor[HTML]{DAE8FC}68.4 & \cellcolor[HTML]{DAE8FC}76.5 & \cellcolor[HTML]{DAE8FC}69.2 & 36.31 \\ \midrule
ILOD \cite{shmelkov2017incremental} & 69.9 & 70.4 & 69.4 & 54.3 & 48 & 68.7 & 78.9 & 68.4 & 45.5 & 58.1 & \cellcolor[HTML]{DAE8FC}59.7 & \cellcolor[HTML]{DAE8FC}72.7 & \cellcolor[HTML]{DAE8FC}73.5 & \cellcolor[HTML]{DAE8FC}73.2 & \cellcolor[HTML]{DAE8FC}66.3 & \cellcolor[HTML]{DAE8FC}29.5 & \cellcolor[HTML]{DAE8FC}63.4 & \cellcolor[HTML]{DAE8FC}61.6 & \cellcolor[HTML]{DAE8FC}69.3 & \cellcolor[HTML]{DAE8FC}62.2 & 63.15 \\
ILOD + Faster R-CNN & 70.5 & 75.6 & 68.9 & 59.1 & 56.6 & 67.6 & 78.6 & 75.4 & 50.3 & 70.8 & \cellcolor[HTML]{DAE8FC}43.2 & \cellcolor[HTML]{DAE8FC}68.1 & \cellcolor[HTML]{DAE8FC}66.2 & \cellcolor[HTML]{DAE8FC}65.1 & \cellcolor[HTML]{DAE8FC}66.5 & \cellcolor[HTML]{DAE8FC}24.3 & \cellcolor[HTML]{DAE8FC}61.3 & \cellcolor[HTML]{DAE8FC}46.6 & \cellcolor[HTML]{DAE8FC}58.1 & \cellcolor[HTML]{DAE8FC}49.9 & 61.14 \\
Faster ILOD~\cite{peng2020faster} & 72.8 & 75.7 & 71.2 & 60.5 & 61.7 & 70.4 & 83.3 & 76.6 & 53.1 & 72.3 & \cellcolor[HTML]{DAE8FC}36.7 & \cellcolor[HTML]{DAE8FC}70.9 & \cellcolor[HTML]{DAE8FC}66.8 & \cellcolor[HTML]{DAE8FC}67.6 & \cellcolor[HTML]{DAE8FC}66.1 & \cellcolor[HTML]{DAE8FC}24.7 & \cellcolor[HTML]{DAE8FC}63.1 & \cellcolor[HTML]{DAE8FC}48.1 & \cellcolor[HTML]{DAE8FC}57.1 & \cellcolor[HTML]{DAE8FC}43.6 & 62.16 \\ 
ORE~\cite{joseph2021towards} & 63.5 & 70.9 & 58.9 & 42.9 & 34.1 & 76.2 & 80.7 & 76.3 & 34.1 & 66.1 & \cellcolor[HTML]{DAE8FC}56.1 & \cellcolor[HTML]{DAE8FC}70.4 & \cellcolor[HTML]{DAE8FC}80.2 & \cellcolor[HTML]{DAE8FC}72.3 & \cellcolor[HTML]{DAE8FC}81.8 & \cellcolor[HTML]{DAE8FC}42.7 & \cellcolor[HTML]{DAE8FC}71.6 & \cellcolor[HTML]{DAE8FC}68.1 & \cellcolor[HTML]{DAE8FC}77 & \cellcolor[HTML]{DAE8FC}67.7 & 64.58 \\ \midrule

ROSETTA-Faster R-CNN & 74.2 & 76.2 & 64.9 & 54.4 & 57.4 & 76.1 & 84.4 & 68.8 & 52.4 & 67.0 & \cellcolor[HTML]{DAE8FC}62.9 & \cellcolor[HTML]{DAE8FC}63.3 & \cellcolor[HTML]{DAE8FC}79.8 & \cellcolor[HTML]{DAE8FC}72.8 & \cellcolor[HTML]{DAE8FC}78.1 & \cellcolor[HTML]{DAE8FC}40.1 & \cellcolor[HTML]{DAE8FC}62.3 & \cellcolor[HTML]{DAE8FC}61.2 & \cellcolor[HTML]{DAE8FC}72.4 & \cellcolor[HTML]{DAE8FC}66.8 & \textbf{66.80}\\
\midrule\midrule

{\color[HTML]{009901} \textbf{15 + 5 setting}} & aero & cycle & bird & boat & bottle & bus & car & cat & chair & cow & table & dog & horse & bike & person & \cellcolor[HTML]{DAE8FC}plant & \cellcolor[HTML]{DAE8FC}sheep & \cellcolor[HTML]{DAE8FC}sofa & \cellcolor[HTML]{DAE8FC}train & \cellcolor[HTML]{DAE8FC}tv & \textbf{mAP} \\ \midrule
First 15 & 74.2 & 79.1 & 71.3 & 60.3 & 60 & 80.2 & 88.1 & 80.2 & 48.8 & 74.6 & 61 & 76 & 85.3 & 78.2 & 83.4 & \cellcolor[HTML]{DAE8FC}0 & \cellcolor[HTML]{DAE8FC}0 & \cellcolor[HTML]{DAE8FC}0 & \cellcolor[HTML]{DAE8FC}0 & \cellcolor[HTML]{DAE8FC}0 & 55.03 \\
New 5 & 3.7 & 0.5 & 6.3 & 4.6 & 0.9 & 0 & 8.8 & 3.9 & 0 & 0.4 & 0 & 0 & 16.4 & 0.7 & 0 & \cellcolor[HTML]{DAE8FC}41 & \cellcolor[HTML]{DAE8FC}55.7 & \cellcolor[HTML]{DAE8FC}49.2 & \cellcolor[HTML]{DAE8FC}59.1 & \cellcolor[HTML]{DAE8FC}67.8 & 15.95 \\ \midrule
ILOD \cite{shmelkov2017incremental} & 70.5 & 79.2 & 68.8 & 59.1 & 53.2 & 75.4 & 79.4 & 78.8 & 46.6 & 59.4 & 59 & 75.8 & 71.8 & 78.6 & 69.6 & \cellcolor[HTML]{DAE8FC}33.7 & \cellcolor[HTML]{DAE8FC}61.5 & \cellcolor[HTML]{DAE8FC}63.1 & \cellcolor[HTML]{DAE8FC}71.7 & \cellcolor[HTML]{DAE8FC}62.2 & 65.87 \\
ILOD + Faster R-CNN & 63.5 & 76.3 & 70.7 & 53.1 & 55.8 & 67.1 & 81.5 & 80.3 & 49.6 & 73.8 & 62.1 & 77.1 & 79.7 & 74.2 & 73.9 & \cellcolor[HTML]{DAE8FC}37.1 & \cellcolor[HTML]{DAE8FC}59.1 & \cellcolor[HTML]{DAE8FC}61.7 & \cellcolor[HTML]{DAE8FC}68.6 & \cellcolor[HTML]{DAE8FC}61.3 & 66.35 \\
Faster ILOD~\cite{peng2020faster} & 66.5 & 78.1 & 71.8 & 54.6 & 61.4 & 68.4 & 82.6 & 82.7 & 52.1 & 74.3 & 63.1 & 78.6 & 80.5 & 78.4 & 80.4 & \cellcolor[HTML]{DAE8FC}36.7 & \cellcolor[HTML]{DAE8FC}61.7 & \cellcolor[HTML]{DAE8FC}59.3 & \cellcolor[HTML]{DAE8FC}67.9 & \cellcolor[HTML]{DAE8FC}59.1 & 67.94 \\ 
ORE~\cite{joseph2021towards} & 75.4 & 81 & 67.1 & 51.9 & 55.7 & 77.2 & 85.6 & 81.7 & 46.1 & 76.2 & 55.4 & 76.7 & 86.2 & 78.5 & 82.1 & \cellcolor[HTML]{DAE8FC}32.8 & \cellcolor[HTML]{DAE8FC}63.6 & \cellcolor[HTML]{DAE8FC}54.7 & \cellcolor[HTML]{DAE8FC}77.7 & \cellcolor[HTML]{DAE8FC}64.6 & 68.51 \\\midrule

ROSETTA-Faster R-CNN & 76.5 & 77.5 & 65.1 & 56.0 & 60.0 & 78.3 & 85.5 & 78.7 & 49.5 & 68.2 & 67.4 & 71.2 & 83.9 & 75.7 & 82.0 & \cellcolor[HTML]{DAE8FC}43.0 & \cellcolor[HTML]{DAE8FC}60.6 & \cellcolor[HTML]{DAE8FC}64.1 & \cellcolor[HTML]{DAE8FC}72.8 & \cellcolor[HTML]{DAE8FC}67.4 & \textbf{69.17}\\

\midrule
\midrule
{\color[HTML]{009901} \textbf{19 + 1 setting}} & aero & cycle & bird & boat & bottle & bus & car & cat & chair & cow & table & dog & horse & bike & person & plant & sheep & sofa & train & \cellcolor[HTML]{DAE8FC}tv & \textbf{mAP} \\ \midrule
First 19 & 77.8 & 81.7 & 69.3 & 51.6 & 55.3 & 74.5 & 86.3 & 80.2 & 49.3 & 82 & 63.6 & 76.8 & 80.9 & 77.5 & 82.4 & 42.9 & 73.9 & 70.4 & 70.4 & \cellcolor[HTML]{DAE8FC}0 & 67.34 \\
Last 1 & 0 & 0 & 0 & 0 & 0 & 0 & 0 & 0 & 0 & 0 & 0 & 0 & 0 & 0 & 0 & 0 & 0 & 0 & 0 & \cellcolor[HTML]{DAE8FC}64 & 3.2 \\ \midrule
ILOD \cite{shmelkov2017incremental} & 69.4 & 79.3 & 69.5 & 57.4 & 45.4 & 78.4 & 79.1 & 80.5 & 45.7 & 76.3 & 64.8 & 77.2 & 80.8 & 77.5 & 70.1 & 42.3 & 67.5 & 64.4 & 76.7 & \cellcolor[HTML]{DAE8FC}62.7 & 68.25 \\
ILOD + Faster R-CNN & 60.9 & 74.6 & 70.8 & 56 & 51.3 & 70.7 & 81.7 & 81.5 & 49.45 & 78.3 & 58.3 & 79.5 & 79.1 & 74.8 & 75.7 & 42.8 & 74.7 & 61.2 & 67.2 & \cellcolor[HTML]{DAE8FC}65.1 & 67.72 \\
Faster ILOD~\cite{peng2020faster} & 64.2 & 74.7 & 73.2 & 55.5 & 53.7 & 70.8 & 82.9 & 82.6 & 51.6 & 79.7 & 58.7 & 78.8 & 81.8 & 75.3 & 77.4 & 43.1 & 73.8 & 61.7 & 69.8 & \cellcolor[HTML]{DAE8FC}61.1 & 68.56 \\

ORE~\cite{joseph2021towards} & 67.3 & 76.8 & 60 & 48.4 & 58.8 & 81.1 & 86.5 & 75.8 & 41.5 & 79.6 & 54.6 & 72.8 & 85.9 & 81.7 & 82.4 & 44.8 & 75.8 & 68.2 & 75.7 & \cellcolor[HTML]{DAE8FC}60.1 & 68.89 \\ \midrule

ROSETTA-Faster R-CNN & 75.3 & 77.9 & 65.3 & 56.2 & 55.3 & 79.6 & 84.6 & 72.9 & 49.2 & 73.7 & 68.3 & 71.0 & 78.9 & 77.7 & 80.7 & 44.0 & 69.6 & 68.5 & 76.1 & \cellcolor[HTML]{DAE8FC}68.3 & \textbf{69.64}\\

\bottomrule
\end{tabular}%
}
\vspace{-3mm}
\caption{Comparisons with existing class-incremental object detectors with Faster R-CNN backbone on three different settings: ``10 + 10", ``15 + 5", ``19 + 1". For example, ``First 15" means training on the first 15 classes of Pascal VOC 2007 with Faster R-CNN backbone and ``New 5" refers to fine-tuning on the new 5 classes. The best results are denoted in boldface.}
\label{tab:class_incre}
\vspace{-15pt}
\end{table*}

\noindent\textbf{Implementation Details} \; Our experiments are conducted on 8 GPUs with batch size of 16 and implemented by PyTorch~\cite{paszke2017automatic}. Without loss of generality, our ROSETTA can be equipped different object detector backbones. In our experiment, we choose two representative object detectors as our backbones: Faster R-CNN~\cite{ren2015faster} and Sparse R-CNN~\cite{sun2021sparse}. Faster R-CNN is a commonly used detector which heavily relies on dense object candidates. Sparse R-CNN is a recently proposed detector without NMS, which has a sparse-in sparse-out paradigm with a higher efficiency. For Faster R-CNN backbone, we use the same training scheme as~\cite{liu2020multi} and generate our baseline results of joint training and fine-tuning. As for Sparse R-CNN, we use its default hyper-parameters with 100 proposal boxes. Specifically, we use $3\times$ training schedule ($36$ epochs) on COCO and VOC and the learning rate is set to $2.5 \times 10^{-5}$ for early training stages, divided by $10$ at epoch $27$ and $33$, respectively. On account of the small sizes of KITTI and Kitchen, we train the Sparse R-CNN for $9$k iterations instead. Our gated module proposed in~\cref{sec:gated_objecter} is applied to the last three stages of the ResNet50~\cite{he2016deep} backbone for both Faster R-CNN and Sparse R-CNN while the first two stages are fixed during training. We first pre-train a non-sparse detector without gated module and use it to guide the training of the gated detector with distillation loss. As mentioned in~\cref{sec:gated_objecter}, our gated model is further fine-tuned for $2$ epochs for better co-adaptation between the binary gates and the channel weights. The Pascal VOC mean average precision~(mAP) is used as our evaluation metric following~\cite{liu2020multi}. 

\vspace{-4pt}
\subsection{Comparison with Existing Methods}
\vspace{-4pt}
For task-incremental object detection, we compare our ROSETTA with AFD~\cite{liu2020multi}, LwF Detection~\cite{shmelkov2017incremental}, Feature Distillation~\cite{romero2014fitnets}, Attention Distillation~\cite{zagoruyko2016paying}, EWC~\cite{kirkpatrick2017overcoming}, MAS~\cite{aljundi2018memory}, the joint training and fine-tuning baselines. In terms of joint training which is usually seen as the empirical upper bound of continual learning, data of all tasks are accessible during training the object detector. As for fine-tuning, all tasks are sequentially trained without additional constraints. For class-incremental object detection, the methods we compare include ILOD~\cite{shmelkov2017incremental}, Faster ILOD~\cite{peng2020faster}, ORE~\cite{joseph2021towards}. The performance of our ROSETTA with Faster R-CNN and Sparse R-CNN backbones~(termed as ``ROSETTA-Faster R-CNN" and ``ROSETTA-Sparse R-CNN" respectively) are given in~\cref{tab:task_incre} and~\cref{tab:class_incre}.

\noindent\textbf{Task-incremental object detection} \; Here we divide task-incremental object detection into two case: $(1)$ domains of tasks are relatively similar, $(2)$ domains are significantly different. These two cases are corresponding to our experiments conducted on COCO-VOC and KITTI-Kitchen, respectively. Results are given in~\cref{tab:task_incre}. Although ROSETTA does not need any exemplar for experience replay, it also outperforms state-of-the-art methods including exemplar-based methods which are denoted by `$\dag$'. Thanks to the gating mechanism, ROSETTA can better obviate the problem of catastrophic forgetting, even in the case of a significant domain gap. Specifically, on Kitchen$\rightarrow$KITTI and KITTI$\rightarrow$Kitchen, the performance of other methods on $\mathtt{task 1}$ dramatically drop while our ROSETTA-Faster R-CNN can alleviate the catastrophic forgetting to a great extent. For example, on KITTI$\rightarrow$Kitchen, ROSETTA-Faster R-CNN achieves 17.3 mAP and 6.1 mAP improvements on KITTI and Kitchen compared to AFD~\cite{liu2020multi}, respectively. Equipped with Sparse R-CNN, ROSETTA-Sparse R-CNN outperforms our fine-tuning baseline on $\mathtt{task 1}$, validating its effectiveness of avoiding catastrophic forgetting.

\noindent\textbf{Class-incremental object detection} \; For the  setting of class-incremental object detection, we mainly compare our ROSETTA with other incremental detectors with a fair Faster R-CNN backbone, in terms of three settings (``10 + 10", ``15 + 5", ``19 + 1") following~\cite{joseph2021towards}.
As shown in~\cref{tab:class_incre}, our ROSETA-Faster R-CNN performs well on the class-incremental tasks against the state-of-the-art incremental detectors~\cite{shmelkov2017incremental,peng2020faster,joseph2021towards}. Notably, ROSETTA-Faster R-CNN surpasses ORE~\cite{joseph2021towards} by 2.2 mAP on the setting of ``10 + 10".

\begin{table}[h]
\vspace{-5pt}
\setlength{\tabcolsep}{4pt}
\centering
\resizebox{\linewidth}{!}{%
\begin{tabular}{l|c|c}
\toprule
& COCO $\rightarrow$ VOC $\rightarrow$ KITTI $\rightarrow$ Kitchen & Average\\
\hline
Fine-tuning & 4.3 \;\,\qquad 16.3 \qquad 45.9 \qquad 87.0  & 38.4 \\
ROSETTA(Ours) & 49.5 \qquad 82.3 \qquad 62.5 \qquad 87.3  & \textbf{70.4}\\
\midrule
& KITTI $\rightarrow$ COCO $\rightarrow$ VOC $\rightarrow$ Kitchen & Average\\
\hline
Fine-tuning & 24.9 \;\,\qquad 5.9 \qquad 24.6 \qquad 85.2  & 35.2 \\
ROSETTA(Ours) & 53.3 \qquad 48.9 \qquad 80.3 \qquad 84.5  & \textbf{66.8}\\
\bottomrule
\end{tabular}}
\vspace{-3mm}
\caption{Results of sequential training on 4 tasks.}
\vspace{-10pt}
\label{tab:sequence}
\end{table}

\begin{table}[h]
\vspace{-5pt}
\centering
\scriptsize
\setlength{\tabcolsep}{1.2pt}
\begin{tabular}{c|c|c|c|c|c|c|c}
    \toprule
     \multirow{2}{*}{$\mathcal{L}_{\textit{sparsity}}$} &
     \multirow{2}{*}{$\mathcal{L}_{\textit{kd}}$}  & \multirow{2}{*}{$\mathcal{L}_{\textit{diversity}}$} &
    \multirow{2}{*}{KITTI $\rightarrow$ Kitchen}
    & \multicolumn{4}{c}{Gates}\\
    \cline{5-8}
    & & & & only $\mathtt{task 1}$ & overlap & only $\mathtt{task 2}$ & not used\\
    \midrule
    \xmark & \xmark & \xmark & 51.7 \qquad 70.2 & 6.8\% & 48.3\% & 9.4\% & 35.5\% \\
    \cmark & \xmark & \xmark & 51.0 \qquad 70.5 & 6.1\% & 26.4\% & 1.2\%  & 66.3\% \\
    \cmark & \cmark & \xmark & 53.3 \qquad 73.3 & 9.6\% & 32.9\% & 8.1\% & 49.4\% \\
    \cmark & \cmark & \cmark & 53.3 \qquad 78.3 & 14.3\% & 28.2\% & 20.3\% & 37.2\% \\
    \bottomrule
\end{tabular}
\vspace{-5pt}
\caption{Ablation studies on KITTI$\rightarrow$Kitchen.}
\label{tab:ablation}
\vspace{-15pt}
\end{table}

\subsection{Results on Four Sequential Tasks}
\vspace{-4pt}
To verify the capability of solving multiple tasks, we compare ROSETTA to the fine-tuning baseline with the same Sparse R-CNN backbone on two task-incremental sequences: COCO $\rightarrow$ VOC $\rightarrow$ KITTI $\rightarrow$ Kitchen and KITTI $\rightarrow$ COCO $\rightarrow$ VOC $\rightarrow$ Kitchen. The sequential results shown in~\cref{tab:sequence} illustrate that our ROSETTA outperforms the fine-tuning strategy by a large margin~(32 mAP and 31.6 mAP in average of four tasks, respectively).

\vspace{-4pt}
\subsection{Ablation Study}
\vspace{-4pt}
\noindent \textbf{The effect of $\mathcal{L}_{\textit{sparsity}}$, $\mathcal{L}_{\textit{kd}}$ and $\mathcal{L}_{\textit{diversity}}$} \; To verify the effectiveness of each module in our proposed ROSETTA, we conduct ablation experiments on KITTI$\rightarrow$Kitchen with Sparse R-CNN backbone to ablate $\mathcal{L}_{\textit{sparsity}}$ $\mathcal{L}_{\textit{kd}}$ and $\mathcal{L}_{\textit{diversity}}$. As shown in~\cref{tab:ablation}, our baseline model,~\emph{i.e.}~the trivial gated detector proposed in \cref{sec:gated_objecter}, can alleviate the catastrophic forgetting while two tasks both occupies lots of channel gates with a great overlap of $48.3\%$. Thanks to the sparsity constraint, we obtain sparser models for both $\mathtt{task1}$ and $\mathtt{task2}$. Moreover, knowledge distillation helps the gated module achieve better convergence due to the guidance of non-gated teacher. In \cref{tab:ablation}, $\mathcal{L}_{\textit{kd}}$ benefits the gate learning of both $\mathtt{task1}$ and $\mathtt{task2}$ and we observe performance improvement on both two tasks. By introducing our proposed task correlation guided gating diversity loss $\mathcal{L}_{\textit{diversity}}$, ROSETTA can automatically capture the significant domain gap between KITTI and Kitchen. Therefore, more exclusive knowledge is exploited for $\mathtt{task2}$  ($20.3\%$ only for $\mathtt{task2}$) and less gates are overlapped by two tasks. With the gating diversity constraint on $\mathtt{task2}$, we observe a remarkable improvement of $5.0$ mAP on Kitchen.

\vspace{-5pt}
\begin{figure}[h]
\begin{center}
\includegraphics[width=0.9\linewidth]{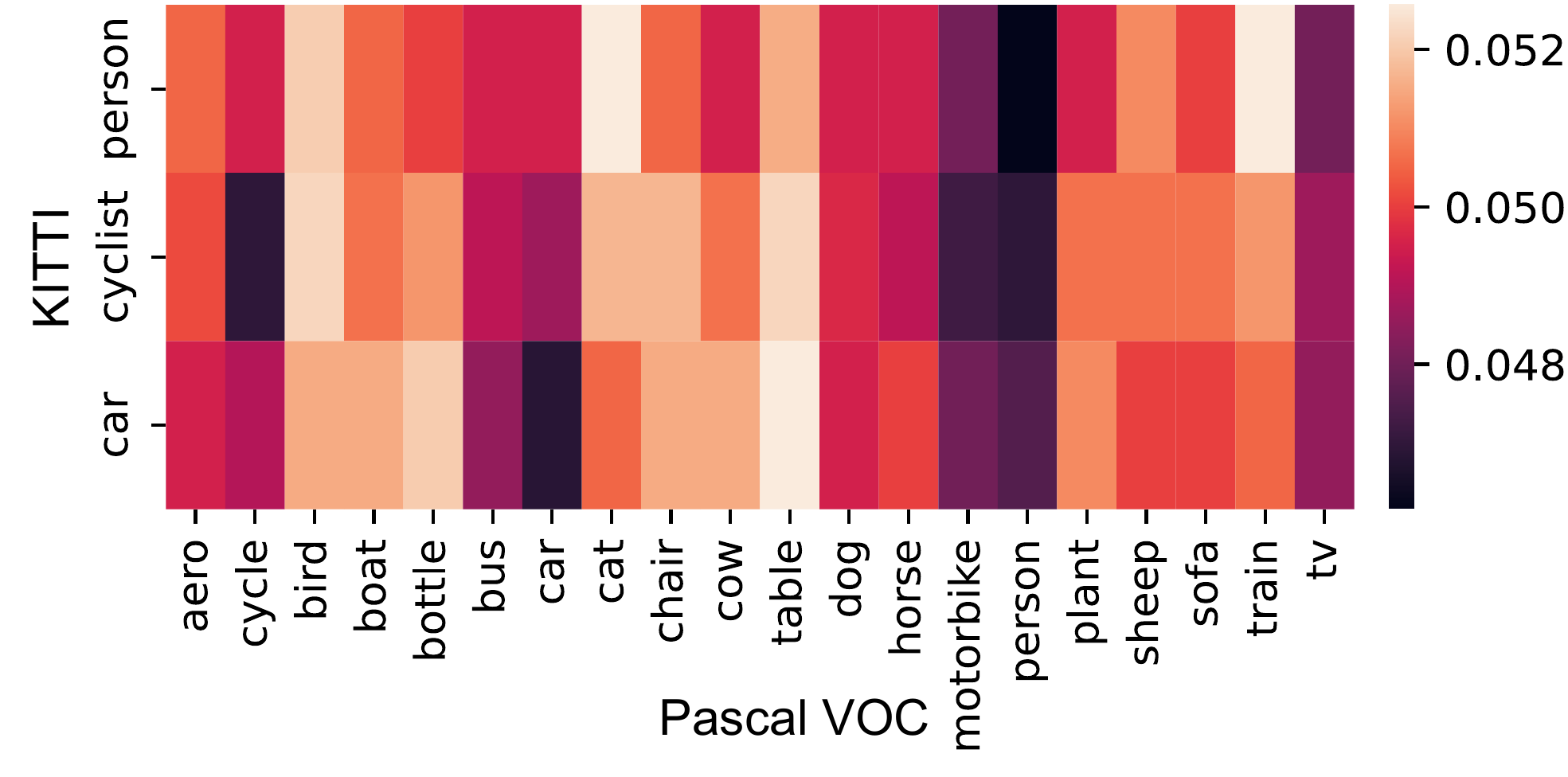}
\end{center}\vspace{-6mm}
   \caption{Visualization of the normalized prototypical correlation matrix of VOC$\rightarrow$KITTI, \emph{i.e.} $\mathbf{M}^{\text{VOC}\rightarrow \text{KITTI}}$. Darker color indicates higher similarity between the classes of these two tasks. \vspace{-15pt}}
\label{fig:prototypical_correlation}
\end{figure}

\subsection{Cross-task Analysis}
\vspace{-4pt}

\noindent \textbf{Cross-task correlation} To understand how ROSETTA captures the cross-task correlation, we visualize the normalized class-to-class prototypical correlation matrix of VOC$\rightarrow$KITTI, \emph{i.e.} $\mathbf{M}^{\text{VOC}\rightarrow \text{KITTI}}$. For convenience, the visualized matrix in~\cref{fig:prototypical_correlation} is its transpose. As illustrated in~\cref{fig:prototypical_correlation}, ROSETTA is able to find out the correct category correspondences between KITTI and VOC. For example, ``cyclist" means people engaged in cycling. Thus, in~\cref{fig:prototypical_correlation}, the class of ``cyclist" in KITTI is close to ``cycle", ``motorbike" which has similar appearance and semantics to ``cycle" and ``person" in VOC.

\begin{figure}[t!]
\vspace{-5pt}
\begin{center}
\includegraphics[width=0.85\linewidth]{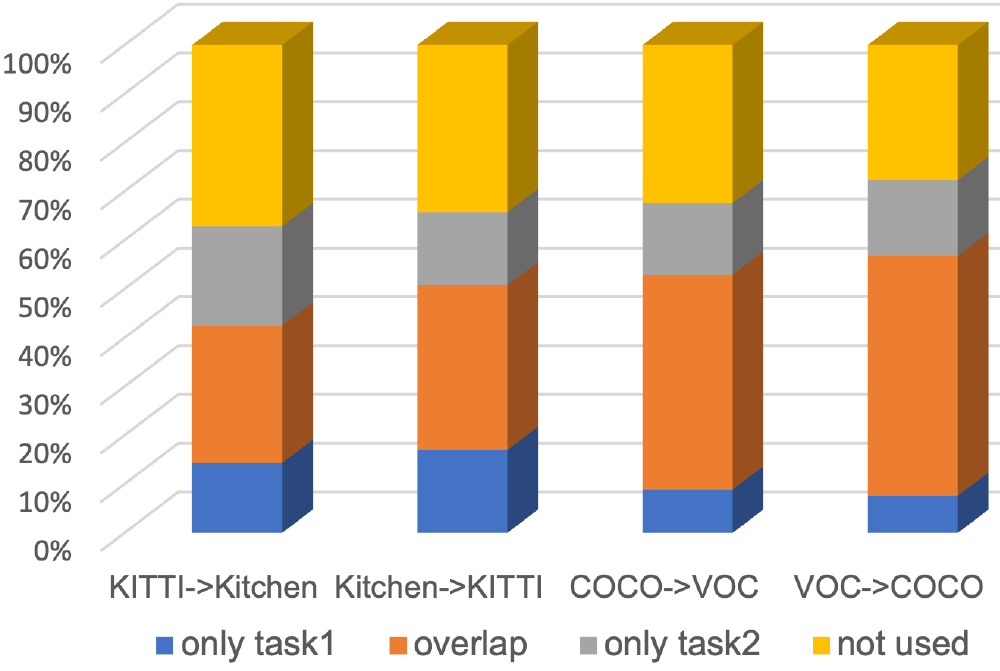}
\end{center}\vspace{-7mm}
   \caption{Analysis of gates for task-incremental detection.\vspace{-15pt}}
\label{fig:gate_analysis}
\end{figure}

\noindent \textbf{Gate analysis}  \; In~\cref{fig:gate_analysis}, we provide the statistical information of the learned binary gates on KITTI-Kitchen and COCO-VOC. Here colors and the corresponding percentage represent the proportion of channel gates activated for different tasks. Since domains of COCO and VOC are similar, the gates learned on $\mathtt{task 1}$ are almost inherited by $\mathtt{task 2}$. By comparison, our 
ROSETTA automatically finds that it needs to activate more channels for $\mathtt{task 2}$ by observing the domain gap between KITTI and Kitchen.

\vspace{-5pt}
\section{Conclusion and Discussion}
\vspace{-5pt}
\label{sec:conclusion}
In this paper, we propose a sparse and dynamic framework for continual object detection via p\textbf{R}ot\textbf{O}typical ta\textbf{S}k corr\textbf{E}la\textbf{T}ion guided ga\textbf{T}ing mech\textbf{A}nism (ROSETTA) to memorize previous knowledge and further promote learning future tasks by knowledge sharing. We propose a task-aware gated module and integrate it with the backbone of object detectors. In this way, the gated detector is capable of avoiding catastrophic forgetting by storing the sub-models' weight and the corresponding binary gates. To further promote learning subsequent tasks, we propose a task correlation guided gating diveristy controller to capture the cross-task correlation. 
Comprehensive experiments on COCO-VOC, KITTI-Kitchen and class-incremental detection on VOC and sequential learning of four tasks verifies the superiority of our ROSETTA on both class-based and task-based continual object detection. 

\noindent \textbf{Potential negative social impact} Our method has no ethical risk on dataset usage and privacy violation as all the benchmarks are public and transparent.

\noindent \textbf{Limitation} With regards to the limitation of our work, we only focus on one kind of task (modality) for the continual learning system~\emph{i.e} object detection in the visual modality. As our expectation, we aim to extend our ROSETTA to a more general continual system which can deal with different kinds of tasks and modalities in future work.

\vspace{-8pt}
\section*{Acknowledgment}
\vspace{-6pt}
This work was supported in part by National Key R\&D Program of China under Grant No. 2020AAA0109700, National Natural Science Foundation of China (NSFC) No.61976233, Guangdong Province Basic and Applied Basic Research (Regional Joint Fund-Key) Grant No.2019B1515120039, Guangdong Outstanding Youth Fund (Grant No. 2021B1515020061).


{\small
\bibliographystyle{ieee_fullname}
\bibliography{egbib}

\begin{thebibliography}{10}\itemsep=-1pt

\bibitem{abati2020conditional}
D. Abati, J. Tomczak, T. Blankevoort, S. Calderara, R. Cucchiara, and B.
  Bejnordi.
\newblock Conditional channel gated networks for task-aware continual learning.
\newblock In {\em CVPR}, 2020.

\bibitem{aljundi2018memory}
R. Aljundi, F. Babiloni, M. Elhoseiny, M. Rohrbach, and T. Tuytelaars.
\newblock Memory aware synapses: Learning what (not) to forget.
\newblock In {\em ECCV}, 2018.

\bibitem{aljundi2018selfless}
R. Aljundi, M. Rohrbach, and T. Tuytelaars.
\newblock Selfless sequential learning.
\newblock In {\em ICLR}, 2019.

\bibitem{bejnordi2019batch}
Babak~Ehteshami Bejnordi, Tijmen Blankevoort, and Max Welling.
\newblock Batch-shaping for learning conditional channel gated networks.
\newblock In {\em ICLR}, 2020.

\bibitem{Bolukbasi2017AdaptiveNN}
Tolga Bolukbasi, Joseph Wang, Ofer Dekel, and Venkatesh Saligrama.
\newblock Adaptive neural networks for fast test-time prediction.
\newblock {\em arXiv:1702.07811}, 2017.

\bibitem{buzzega2020dark}
Pietro Buzzega, Matteo Boschini, Angelo Porrello, Davide Abati, and Simone
  Calderara.
\newblock Dark experience for general continual learning: a strong, simple
  baseline.
\newblock In {\em 34th Conference on Neural Information Processing Systems
  (NeurIPS 2020)}, 2020.

\bibitem{chaudhry2018efficient}
A. Chaudhry, M. Ranzato, M. Rohrbach, and M. Elhoseiny.
\newblock Efficient lifelong learning with a-gem.
\newblock In {\em ICLR}, 2019.

\bibitem{chen2019self}
Jinting Chen, Zhaocheng Zhu, Cheng Li, and Yuming Zhao.
\newblock Self-adaptive network pruning.
\newblock In {\em ICONIP}, 2019.

\bibitem{chen2019progressive}
Xin Chen, Lingxi Xie, Jun Wu, and Qi Tian.
\newblock Progressive differentiable architecture search: Bridging the depth
  gap between search and evaluation.
\newblock In {\em Proceedings of the IEEE/CVF International Conference on
  Computer Vision}, pages 1294--1303, 2019.

\bibitem{Chen2019YouLT}
Zhourong Chen, Yang Li, Samy Bengio, and Si Si.
\newblock You look twice: Gaternet for dynamic filter selection in cnns.
\newblock In {\em {CVPR}}, 2019.

\bibitem{everingham2010pascal}
M. Everingham, L. Van~Gool, C. Williams, J. Winn, and A. Zisserman.
\newblock The pascal visual object classes (voc) challenge.
\newblock {\em International journal of computer vision}, 2010.

\bibitem{french1999catastrophic}
R. French.
\newblock Catastrophic forgetting in connectionist networks.
\newblock {\em Trends in Cognitive Sciences}, 1999.

\bibitem{gao2018dynamic}
Xitong Gao, Yiren Zhao, {\L}ukasz Dudziak, Robert Mullins, and Cheng-zhong Xu.
\newblock Dynamic channel pruning: Feature boosting and suppression.
\newblock In {\em International Conference on Learning Representations}, 2018.

\bibitem{geiger2012we}
A. Geiger, P. Lenz, and R. Urtasun.
\newblock Are we ready for autonomous driving? the kitti vision benchmark
  suite.
\newblock In {\em CVPR}, 2012.

\bibitem{georgakis2016multiview}
G. Georgakis, M. Reza, A. Mousavian, P. Le, and J. Ko{\v{s}}eck{\'a}.
\newblock Multiview rgb-d dataset for object instance detection.
\newblock In {\em International Conference on 3D Vision}, 2016.

\bibitem{guo2020single}
Zichao Guo, Xiangyu Zhang, Haoyuan Mu, Wen Heng, Zechun Liu, Yichen Wei, and
  Jian Sun.
\newblock Single path one-shot neural architecture search with uniform
  sampling.
\newblock In {\em European Conference on Computer Vision}, pages 544--560.
  Springer, 2020.

\bibitem{hadsell2020embracing}
R. Hadsell, D. Rao, A. Rusu, and R. Pascanu.
\newblock Embracing change: Continual learning in deep neural networks.
\newblock {\em Trends in Cognitive Sciences}, 2020.

\bibitem{he2016deep}
K. He, X. Zhang, S. Ren, and J. Sun.
\newblock Deep residual learning for image recognition.
\newblock In {\em CVPR}, 2016.

\bibitem{herrmann2018end}
Charles Herrmann, Richard~Strong Bowen, and Ramin Zabih.
\newblock An end-to-end approach for speeding up neural network inference.
\newblock {\em arXiv preprint arXiv:1812.04180}, 2018.

\bibitem{hinton2015distilling}
Geoffrey Hinton, Oriol Vinyals, and Jeff Dean.
\newblock Distilling the knowledge in a neural network.
\newblock {\em stat}, 1050:9, 2015.

\bibitem{hua2019channel}
Weizhe Hua, Yuan Zhou, Christopher~M De~Sa, Zhiru Zhang, and G~Edward Suh.
\newblock Channel gating neural networks.
\newblock In {\em NeurIPS}, 2019.

\bibitem{Huang2018MultiScaleDN}
Gao Huang, Danlu Chen, Tianhong Li, Felix Wu, Laurens van~der Maaten, and
  Kilian~Q. Weinberger.
\newblock Multi-scale dense networks for resource efficient image
  classification.
\newblock In {\em ICLR}, 2018.

\bibitem{joseph2021towards}
KJ Joseph, Salman Khan, Fahad~Shahbaz Khan, and Vineeth~N Balasubramanian.
\newblock Towards open world object detection.
\newblock In {\em Proceedings of the IEEE/CVF Conference on Computer Vision and
  Pattern Recognition}, pages 5830--5840, 2021.

\bibitem{kirkpatrick2017overcoming}
J. Kirkpatrick, R. Pascanu, N. Rabinowitz, J. Veness, G. Desjardins, A. Rusu,
  K. Milan, J. Quan, T. Ramalho, A. Grabska-Barwinska, et~al.
\newblock Overcoming catastrophic forgetting in neural networks.
\newblock {\em National Academy of Sciences}, 2017.

\bibitem{Li2021DynamicSN}
Changlin L., Guangrun W., Bing W., Xiaodan L., Zhihui L., and Xiaojun C.
\newblock Dynamic slimmable network.
\newblock In {\em CVPR}, 2021.

\bibitem{lee2021sharing}
Seungwon Lee, Sima Behpour, and Eric Eaton.
\newblock Sharing less is more: Lifelong learning in deep networks with
  selective layer transfer.
\newblock In {\em International Conference on Machine Learning}, pages
  6065--6075. PMLR, 2021.

\bibitem{li2020DynamicRouting}
Yanwei Li, Lin Song, Yukang Chen, Zeming Li, Xiangyu Zhang, Xingang Wang, and
  Jian Sun.
\newblock Learning dynamic routing for semantic segmentation.
\newblock In {\em CVPR}, 2020.

\bibitem{li2017learning}
Z. Li and D. Hoiem.
\newblock Learning without forgetting.
\newblock In {\em ECCV}, 2016.

\bibitem{lin2017runtime}
Ji Lin, Yongming Rao, Jiwen Lu, and Jie Zhou.
\newblock Runtime neural pruning.
\newblock In {\em NeurIPS}, 2017.

\bibitem{lin2014microsoft}
T. Lin, M. Maire, S. Belongie, J. Hays, P. Perona, D. Ramanan, P. Doll{\'a}r,
  and C. Zitnick.
\newblock Microsoft coco: Common objects in context.
\newblock In {\em ECCV}, 2014.

\bibitem{lin2017feature}
Tsung-Yi Lin, Piotr Doll{\'a}r, Ross Girshick, Kaiming He, Bharath Hariharan,
  and Serge Belongie.
\newblock Feature pyramid networks for object detection.
\newblock In {\em CVPR}, pages 2117--2125, 2017.

\bibitem{lin2017focal}
Tsung-Yi Lin, Priya Goyal, Ross Girshick, Kaiming He, and Piotr Doll{\'a}r.
\newblock Focal loss for dense object detection.
\newblock In {\em Proceedings of the IEEE international conference on computer
  vision}, pages 2980--2988, 2017.

\bibitem{liu2018darts}
Hanxiao Liu, Karen Simonyan, and Yiming Yang.
\newblock Darts: Differentiable architecture search.
\newblock In {\em International Conference on Learning Representations}, 2018.

\bibitem{liu2020multi}
X. Liu, H. Yang, A. Ravichandran, R. Bhotika, and S. Soatto.
\newblock Multi-task incremental learning for object detection.
\newblock {\em arXiv:2002.05347}, 2020.

\bibitem{lopez2017gradient}
D. Lopez-Paz and M. Ranzato.
\newblock Gradient episodic memory for continual learning.
\newblock In {\em NeurIPS}, 2017.

\bibitem{mallya2018packnet}
A. Mallya and S. Lazebnik.
\newblock Packnet: Adding multiple tasks to a single network by iterative
  pruning.
\newblock In {\em CVPR}, 2018.

\bibitem{mikolov2013efficient}
Tomas Mikolov, Kai Chen, Greg Corrado, and Jeffrey Dean.
\newblock Efficient estimation of word representations in vector space.
\newblock {\em arXiv preprint arXiv:1301.3781}, 2013.

\bibitem{paszke2017automatic}
A. Paszke, S. Gross, S. Chintala, G. Chanan, E. Yang, Z. DeVito, Z. Lin, A.
  Desmaison, L. Antiga, and A. Lerer.
\newblock Automatic differentiation in pytorch.
\newblock In {\em NIPS Autodiff Workshop}, 2017.

\bibitem{peng2020faster}
C. Peng, K. Zhao, and B. Lovell.
\newblock Faster ilod: Incremental learning for object detectors based on
  faster rcnn.
\newblock {\em Pattern Recognition Letters}, 2020.

\bibitem{rajasegaran2019random}
Jathushan Rajasegaran, Munawar Hayat, Salman Khan, Fahad~Shahbaz Khan, and Ling
  Shao.
\newblock Random path selection for incremental learning.
\newblock {\em Advances in Neural Information Processing Systems}, 2019.

\bibitem{ren2015faster}
S. Ren, K. He, R. Girshick, and J. Sun.
\newblock Faster r-cnn: Towards real-time object detection with region proposal
  networks.
\newblock In {\em NeurIPS}, 2015.

\bibitem{riemer2018learning}
M. Riemer, I. Cases, R. Ajemian, M. Liu, I. Rish, Y. Tu, and G. Tesauro.
\newblock Learning to learn without forgetting by maximizing transfer and
  minimizing interference.
\newblock In {\em ICLR}, 2018.

\bibitem{romero2014fitnets}
A. Romero, N. Ballas, S. Kahou, A. Chassang, C. Gatta, and Y. Bengio.
\newblock Fitnets: Hints for thin deep nets.
\newblock {\em arXiv:1412.6550}, 2014.

\bibitem{serra2018overcoming}
J. Serra, D. Suris, M. Miron, and A. Karatzoglou.
\newblock Overcoming catastrophic forgetting with hard attention to the task.
\newblock In {\em ICML}, 2018.

\bibitem{shmelkov2017incremental}
K. Shmelkov, C. Schmid, and K. Alahari.
\newblock Incremental learning of object detectors without catastrophic
  forgetting.
\newblock In {\em ICCV}, 2017.

\bibitem{sun2021sparse}
Peize Sun, Rufeng Zhang, Yi Jiang, Tao Kong, Chenfeng Xu, Wei Zhan, Masayoshi
  Tomizuka, Lei Li, Zehuan Yuan, Changhu Wang, et~al.
\newblock Sparse r-cnn: End-to-end object detection with learnable proposals.
\newblock In {\em Proceedings of the IEEE/CVF Conference on Computer Vision and
  Pattern Recognition}, pages 14454--14463, 2021.

\bibitem{veit2018AIG}
Andreas Veit and Serge Belongie.
\newblock Convolutional networks with adaptive inference graphs.
\newblock In {\em ECCV}, 2018.

\bibitem{wang2018skipnet}
Xin Wang, Fisher Yu, Zi-Yi Dou, Trevor Darrell, and Joseph~E Gonzalez.
\newblock Skipnet: Learning dynamic routing in convolutional networks.
\newblock In {\em ECCV}, 2018.

\bibitem{wang_dual_2020}
Yue Wang, Jianghao Shen, Ting-Kuei Hu, Pengfei Xu, Tan Nguyen, Richard~G.
  Baraniuk, Zhangyang Wang, and Yingyan Lin.
\newblock Dual dynamic inference: {Enabling} more efficient, adaptive and
  controllable deep inference.
\newblock {\em JSTSP}, 2020.

\bibitem{xia2021fully}
Wenhan Xia, Hongxu Yin, Xiaoliang Dai, and Niraj~K Jha.
\newblock Fully dynamic inference with deep neural networks.
\newblock {\em IEEE Transactions on Emerging Topics in Computing}, 2021.

\bibitem{xu2019pc}
Yuhui Xu, Lingxi Xie, Xiaopeng Zhang, Xin Chen, Guo-Jun Qi, Qi Tian, and
  Hongkai Xiong.
\newblock Pc-darts: Partial channel connections for memory-efficient
  architecture search.
\newblock {\em arXiv preprint arXiv:1907.05737}, 2019.

\bibitem{zagoruyko2016paying}
S. Zagoruyko and N. Komodakis.
\newblock Paying more attention to attention: Improving the performance of
  convolutional neural networks via attention transfer.
\newblock In {\em ICLR}, 2017.

\bibitem{zhou2020lifelong}
Wang Zhou, Shiyu Chang, Norma Sosa, Hendrik Hamann, and David Cox.
\newblock Lifelong object detection.
\newblock {\em arXiv preprint arXiv:2009.01129}, 2020.

\end{thebibliography}
}
\end{document}